\documentclass[conference]{IEEEtran}
\usepackage{cite}
\usepackage{epsfig} 
\usepackage{amsmath,amssymb,amsfonts}
\usepackage{algorithmic}
\usepackage{graphicx}
\usepackage{textcomp}
\usepackage{hyperref}
\usepackage{makecell}
\usepackage{multicol}
\usepackage{xcolor}
\usepackage{multirow}
\usepackage{float}
\usepackage{subcaption}
\def\BibTeX{{\rm B\kern-.05em{\sc i\kern-.025em b}\kern-.08em
    T\kern-.1667em\lower.7ex\hbox{E}\kern-.125emX}}
\begin{document}

\title{{Flickr-PAD: New Face High-Resolution Presentation Attack Detection Database}
\thanks{This work is supported by the European Union’s Horizon 2020 research and innovation program under grant agreement No 883356 and the German Federal Ministry of Education and Research and the Hessian Ministry of Higher Education, Research, Science and the Arts within their joint support of the National Research Center for Applied Cybersecurity ATHENE and TOC Biometric-R\&D Center.}
}

\author{\IEEEauthorblockN{ Diego~Pasmino, Carlos~Aravena}
\IEEEauthorblockA{\textit{R\&D Center, TOC Biometrics} \\
Santiago, Chile. \\
{diego.pasmino, carlos.aravena}@tocbiometrics.com}
\and
\IEEEauthorblockN{ Juan~E.~Tapia, Christoph Busch}
\IEEEauthorblockA{\textit{da/sec-Biometrics and Internet Security Research Group, } \\
\textit{Hochschule Darmstadt, Germany}\\
juan.tapia-farias, christoph.busch@h-da.de}
}

\IEEEoverridecommandlockouts
\IEEEpubid{\makebox[\columnwidth]{979-8-3503-3607-8/23/\$31.00 ©2023 IEEE \hfill} \hspace{\columnsep}\makebox[\columnwidth]{ }}

\maketitle

\IEEEpubidadjcol

\begin{abstract}
Nowadays, Presentation Attack Detection is a very active research area. Several databases are constituted in the state-of-the-art using images extracted from videos. One of the main problems identified is that many databases present a low-quality, small image size and do not represent an operational scenario in a real remote biometric system. Currently, these images are captured from smartphones with high-quality and bigger resolutions. In order to increase the diversity of image quality, this work presents a new PAD database based on open-access Flickr images called: "Flickr-PAD". Our new hand-made database shows high-quality printed and screen scenarios. This will help researchers to compare new approaches to existing algorithms on a wider database. This database will be available for other researchers. A leave-one-out protocol was used to train and evaluate three PAD models based on MobileNet-V3 (small and large) and EfficientNet-B0. The best result was reached with MobileNet-V3 large with BPCER10 of 7.08\% and BPCER20 of 11.15\%. 
\end{abstract}


\section{Introduction}
\label{sec:introduction}
\IEEEPARstart{F}{ace} based biometric systems have numerous commercial and industrial applications in fields as diverse as access controls, video surveillance, and user validation. Due to the improvement in the quality, availability of capture devices, the raising of remote verification systems and transmission capabilities, these applications are increasingly available to the general public in uncontrolled environments. 
In this scenario, one of the most relevant challenges to the viability of biometric systems is the problem of facial impersonation attacks. In this case, a subject presents false evidence (attack)to the capture device or facial recognition system to be authorised and gain access or obtain resources. For example, the security system could be fooled if an unauthorised subject presents in front of the capture device with a high-definition image of the correct user's face. This problem broadly affects end users, companies, governments and others.

Presentation Attack Detection (PAD) is applied to overcome this challenge. This technique consists of differentiating between a pristine biometric capturing of a living subject (\emph{bona fide presentation}) and a fake one created by an attacker (\emph{attack presentation}), using, for example, a photo, video, mask or a different substitute for the face of an authorised subject. 
As a sub-set of the PAD mechanisms, \emph{liveness detection} is used to indicate the act of verifying vitality, such as the pulse or blood flow of the subject presented in front of a capture device.

At present, there are various techniques to solve this challenge.
However, active research is conducted on this topic due to the difficulty involved in the designing an algorithm that generalises well to different capture devices and environmental conditions. 

As previously noted, facial recognition systems are attacked
in several ways, including attacks through printed media, video playback,
use of 2D or 3D masks, etc., specially designed to impersonate an
authorised user. These attacks can be divided into two broad  categories:
digital manipulation attacks, which range from manipulation of direct
information from the capture device or transmitted data, and Presentation Attacks (PA) that try to fool Facial Recognition Systems (FRS) by showing the representation of the target face on some physical media in front of the image capture device. PA tend to be the most common way to attack FRS, especially in uncontrolled or remote environments, because they do not necessarily require too much expertise or knowledge on the part of the attacker trying to be authorised. In summary, the main contributions of this paper are:

\begin{itemize}
    \item  A comprehensive analysis of PAD regarding the effective image sizes is presented.
    \item A new PA database is proposed called "Flickr-PAD" with high-quality images and two hand-made new scenarios such as printed and screen. This database will be available to other researchers by request \footnote{\url{https://github.com/jedota/Flickr-PAD}}.
    \item An exhaustive evaluation protocol was proposed based on intra/cross and leave-one-out protocols considering state-of-the-art databases.
   
\end{itemize}

The rest of the article is organised as follows: Section~\ref{sec:relate} summarises the related works on PAD. The database description is explained in Section~\ref{sec:database}. The metrics are explained in Section~\ref{sec:metric}. The experiment and results framework is then presented in Section~\ref{sec:exp_results}. We conclude the article in Section~\ref{sec:conclusions}.

\section{RELATED WORK}
\label{sec:relate}

Previous PAD image databases provide a variety of attack types under different setups. Several aggregate databases have been created and are described in the literature \cite{yu2022deep}. Despite these databases being mostly video recordings, they are widely used for PAD mechanisms training and testing. This gives the model a high number of images (frames) to be trained with at the cost of losing diversity because the images tend to be similar if selected within a short time from the same video source. 

Deep Learning (DL) has proven to be a great technique to improve the performance of several pattern recognition solutions, given that a sufficiently large amount of data is available. Facial PAD is no exception, and several methods have been proposed based on DL and, more specifically, convolutional networks. A recent review of this kind of algorithm's state of the art is given in \cite{yu2022deep}.

A comprehensive DL-based algorithm is presented in \cite{8987257} for detecting both digital and physical PA, using Cross Asymmetric Loss Function (CALF). The evaluation with a cross-attack / cross-database protocol showed promising results when using the CASIA-MFSD database for training.

A hybrid approach, combining handcrafted Local Binary Patterns (LBP) on the brightness and chrominance channels and DL network features from a VGG-16 pre-trained model, is proposed in \cite{8955089}. The author reported good results when detecting 2D attacks with high accuracy when tested with the Replay-Mobile database.

An illumination-invariant method for PAD is presented in
\cite{8737949}. In this method, a Two-Stream Convolutional Neural
Network (TSCNN), working on RGB space and Multi-Scale Retinex (MSR)
space (illumination-invariant space) is used. MSR images can effectively capture high-frequency information, which is discriminative for face spoofing detection. An attention-based fusion method effectively captures the complementarity of these two features. The CASIA \cite{CASIA-MFSD}, OULU-NPU \cite{OULU-NPU}, and Replay-Attack \cite{replay-attack} databases were used to evaluate the proposed method's performance. Both CASIA and Replay-Attack databases achieved similar performance when used to train and cross-test the model.

An end-to-end Single-Side domain generalisation framework (SSDG) \cite{Jia_2020_CVPR_SSDG} is used to learn a generalised feature space where the feature distribution of the real faces is compact. In contrast, the fake ones are dispersed among domains. Specifically, a feature generator is trained to make only the bona fide faces from different domains undistinguishable but not the fake ones. Feature and weight normalisation are incorporated to improve the generalisation ability further. Experiments show very good results, comparably better than other state-of-the-art methods. The OULU, CASIA, MSU, and Replay databases were used for training and evaluating this method following the leave-one-out protocol.

A dual-stream convolution neural networks (CNNs) framework is proposed
in \cite{fang2021learnable}. One stream adapts four learnable frequency filters to learn features in the frequency domain. The other uses RGB images, which complement the features of the frequency domain. A hierarchical attention module is integrated to join the information from the two streams. This method is evaluated in the intra-dataset and cross-dataset protocols showing good generalisation capabilities in comparison to state-of-the-art. The OULU, CASIA, MSU, and Replay databases were again used for training and evaluating this method following the leave-one-out protocol.

Very recently, \cite{Ebihara} proposed a PAD method based on images captured in-house with and without flash presence. This PAD algorithm used a flash light's specular and diffuse reflection. The iris regions and the facial surface are also used to compute the Speculum Descriptor and the Diffusion Descriptor, respectively. The two descriptors are vectorised and concatenated to build the SpecDiff descriptor, which can be classified as a bona fide or attack by using a standard classifier such as a Support Vector Machine. The four public databases, NUA, Replay-Attack, SiW (test subset), and OULU-NPU (test subset), were used. As we noted in Table \ref{databases_description}, most databases present low-resolution images.

\begin{table}[H]
\centering
\scriptsize
\caption{Public state-of-the-art available databases description. B/A: Bona fide/Attack. P/S: Print/Screen.}
\label{databases_description}
\begin{tabular}{|c|c|c|c|c|}
\hline
Dataset                                                 & B/A          & Subjects & PAI & Resolution                                                          \\ \hline
CASIA-MFSD\cite{CASIA-MFSD}                                              & 150/450      & 50       & P/S & \begin{tabular}[c]{@{}c@{}}$640\times480$\\ $720\times1280$\end{tabular}
\\ \hline
MSU-MFSD\cite{MSU-MFSD}                                                & 70/120       & 35       & P/S & \begin{tabular}[c]{@{}c@{}}$640\times480$\\ $720\times240$\end{tabular}           \\ \hline
OULU-NPU\cite{OULU-NPU}                                                & 720/2,880    & 55       & P/S & $1,920\times1,080$                                                         \\ \hline
\begin{tabular}[c]{@{}c@{}}REPLAY\\ MOBILE \cite{replay-mobile}\end{tabular} & 390/440      & 40       & P/S & $720\times1,280$                                                           \\ \hline
\textbf{Flickr-PAD}                                              & 3,000/11,000 & 3,000    & P/S & \begin{tabular}[c]{@{}c@{}}$6K \times 8K$\\ $3K \times 4K$\\ $4K \times 5K$\end{tabular} \\ \hline
\end{tabular}
\end{table}

In order to improve the previous limitation, a new high-quality database reflecting the capture conditions of a real remote verification system, therefore a new PAD dataset, is proposed in \ref{sec:database_flickr}

\begin{figure*}
\begin{centering}
\includegraphics[scale=0.24]{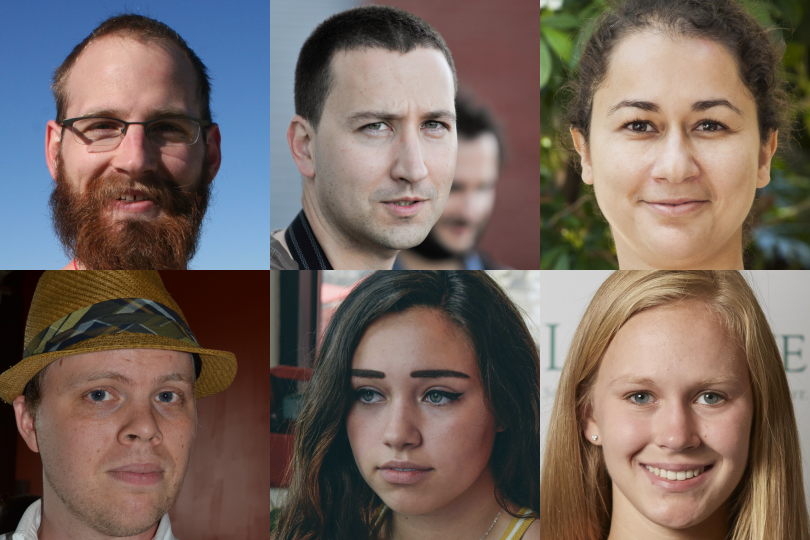}
\includegraphics[scale=0.24]{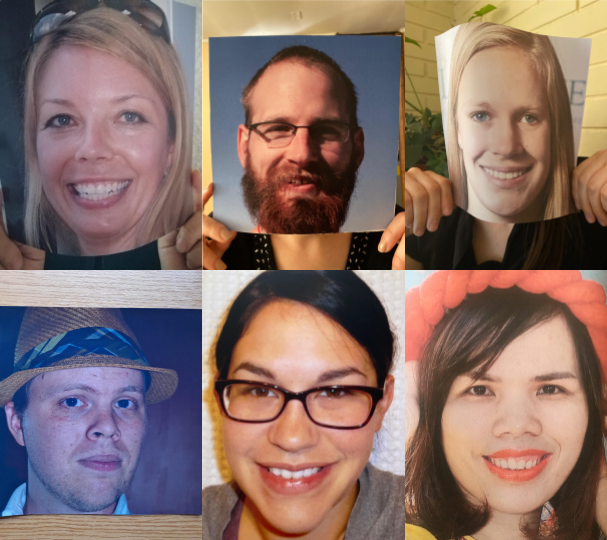}
\includegraphics[scale=0.24]{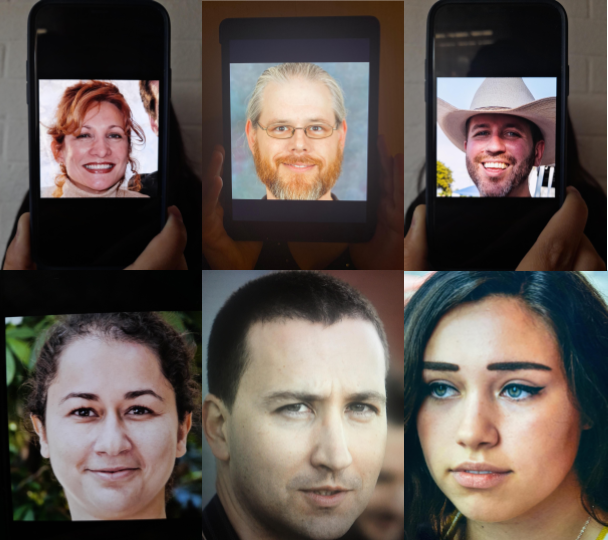}
\caption{\label{Flickr-PAD-images-attacks} Examples of the Flickr-PAD database. Left: bona fide images. Middle: printed attacks. Right: screen attacks.}
\par\end{centering}
\end{figure*}

\section{Datasets}
\label{sec:database}

\subsection{Flickr-PAD Database (F)}
\label{sec:database_flickr}
The new database presented in this work was built based on the Flickr-Face-HQ (FFHQ) Dataset of high-quality images of human faces introduced in \cite{FlickrPaper}.

The individual images were published on Flickr by their respective authors under either Creative Commons BY 2.0, Creative Commons BY-NC 2.0, Public Domain Mark 1.0, Public Domain CC0 1.0, or U.S. Government Works license. All of these licenses allow free use, redistribution, and adaptation for non-commercial purposes \footnote{\url{https://github.com/NVlabs/ffhq-dataset}}. 3,000 face images resolution of $1,024\times1,024$ pixels- from the FFHQ-Dataset were used as bona fide. Only portrait and selfie-like photos were chosen, with clear facial biometric characteristics showing open eyes and a full mouth. Furthermore, facial images showing other objects, such as hats or glasses, were included when they did not interfere with most of the face. A few examples of the selected images are shown in Figure \ref{Flickr-PAD-images-attacks}.

The PA set was made from bona fide images. We divided the 3,000 bona fide images into three groups of 1,000 images. Each group was manually selected and assigned a different Presentation Attack Instrument (PAI) as follows: paper matte, glossy, and bond. Different types of screen monitors (laptops, TVs), smartphones (IPhone-XI, LG, Huawey), and tablets (IPad and Micr. Surf) were also used. Different surface types of paper (i.e. matte or glossy) and printer types (HP-M479 and Epson-2711) were used. The first group is composed of matte paper and a monitor screen. The second group was composed of glossy paper and a smartphone screen; the third group was assigned bond paper and a tablet screen. Two rounds for capturing the PA images were created, and the protocols are described as follows:

\begin{itemize}
    \item The first round consisted of printing the bona fide image on the assigned paper, cutting the borders, placing the print sheet over a flat surface, and then capturing the printed attack with an available smartphone camera (IOS/Android). For the screen attacks, the bona fide image was displayed on the assigned screen, placing the screen device over a flat and steady surface and capturing the attack with a capture device.
    \item The second round consisted of using the already printed faces but holding the sheet with both hands in front of the face, slightly curved, to simulate a limited extent of 3D depth information. For the screen attacks, the assistant held to keep the screen device in front of their face with both hands, displaying the bona fide image to capture the attack with a camera.
\end{itemize}

The Flickr-PAD dataset comprises 3,000 bona fide plus 6,000 printed and 5,000 screen-attacked raw images. The new attack images have the resolution given by the smartphone camera used to capture each attack. A detailed description of the Flickr-PAD dataset is contained in Table \ref{flicker-pad-description}, with the groups of bona fide images, print and screen attacks, and the tools we used to capture the images. Examples of the attack images are shown in Figure \ref{Flickr-PAD-images-attacks}. For the experiments in this paper, the images were pre-processed with the MTCNN face detection algorithm and then cropped to reduce the image size to $256\times256$ pixels.

\subsection{OULU-NPU Database (O)}
The OULU-NPU \cite{OULU-NPU} database contains 2,880 videos of printed and screen face attack presentation samples and is one of the largest sets of videos available. Still, the length of the videos was limited to five seconds. Then, only 55 subjects were recorded. We extracted between 2 to 3 frames from each video in the OULU NLP database, constituting a total of 9,417 images (1,890 bona fide plus 7,527 attacks).
\vspace{-0.3cm}

\subsection{MSU-MFSD Database (M)}
The MSU-MFSD \cite{MSU-MFSD} database has 210 videos of printed and screen face attacks recorded on 35 subjects. The videos have an average length of 12 seconds and 30 fps. In total, we extracted between 2 to 3 frames from each video in the MSU-MFSD Database, collecting in summary 420 images (105 bona fide plus 315 attacks).

\subsection{Replay-Mobile Database (R)}
The REPLAY-ATTACK face spoofing database \cite{replay-attack} includes 1,300 videos from 50 subjects. 100 pristine videos are used for enrolment data for face-verification experiments. The remaining 1,200 are divided into three non-overlapping subsets. These subsets constitute a protocol for unbiased training, tuning and testing new algorithms. We extracted between 9 to 10 frames from each video in the train and development folders of the Replay-Mobile Database, collecting a total of 7,140 images (2,726 bona fide plus 4,414 attacks).

\subsection{CASIA Face Antispoofing Database (C)}
CASIA-MFSD \cite{CASIA-MFSD} is a dataset for face anti-spoofing. It contains 50 subjects and 12 videos for each subject under different resolutions and light conditions. Three different spoof attacks are designed: replay, warp print and cut print attacks. The database contains 600 video recordings, of which 240 videos of 20 subjects are used for training and 360 videos of 30 subjects for testing. In our case, we extracted between 4 to 5 frames from each video in the train folder of the CASIA-Face Antispoofing database, collecting a total of 1,176 images (279 bona fide plus 897 attacks).

The composition of the datasets we used in this work is shown in Table \ref{dataset_composition}.

\begin{table}[H]
\scriptsize
\caption{Proposed database Flickr-PAD description.}
\label{flicker-pad-description}
\centering{}%
\begin{tabular}{|c|c|}
\hline
PAI & Print, Screen\tabularnewline
\hline
Paper type & Matte, Glossy, Regular\tabularnewline
\hline
Screen type & Monitor (PC), Smartphone, Tablet\tabularnewline
\hline
Paper deformation& Flat on a table, Curved in front of assistant\textquoteright s face\tabularnewline
\hline
Screen support& Steady on a table, Held in front of assistant\textquoteright s face\tabularnewline
\hline
\#Bona fide & 3,000\tabularnewline
\hline
\#Print attacks & 6,000\tabularnewline
\hline
\#Screen attacks & 5,000\tabularnewline
\hline
Resolution & \makecell{$1,024\times1,024$ / $6,000\times8,000$ / $3,024\times4,032$ /  \\ $3,000\times4,000$ / $3,840\times5,120$}\tabularnewline
\hline
Scenarios & \makecell{Daylight (morning, afternoon), inside and outside, \\ dark room, artificial light}\tabularnewline
\hline
Capturing devices& iOS and Android\tabularnewline
\hline
\end{tabular}
\end{table}

\begin{table}[H]
\scriptsize
\caption{Dataset composition per PAI.}
\label{dataset_composition}
\centering{}%
\begin{tabular}{|c|c|c|c|c|c|}
\hline
Dataset & Bona fide & Print & Screen & Comments\tabularnewline
\hline
\hline
CASIA-MFSD\cite{CASIA-MFSD} & 279 & 576 & 321 & \makecell{4-5 frames\\per video}\tabularnewline
\hline
MSU-MFSD\cite{MSU-MFSD} & 105 & 105 & 210 & \makecell{2-3 frames\\per video}\tabularnewline
\hline
OULU-NPU\cite{OULU-NPU} & 1,890 & 3,780 & 3,747 &  \makecell{2-3 frames\\per video}\tabularnewline
\hline
Replay-Mobile\cite{replay-mobile} & 2,726 & 3,295 & 1,119 & \makecell{9-10 frames\\per video}\tabularnewline
\hline
Flickr-PAD (\textbf{ours}) & 3,000 & 6,000 & 5,000 & - \tabularnewline
\hline
Total & 9,579 & 15,336 & 11,976 & - \tabularnewline
\hline
\end{tabular}
\end{table}

\section{METRICS}
\label{sec:metric}

PAD systems can be attacked by a potentially large and indeterminate number and diversity of PAI. A PAI species is defined as a class of presentation attack instruments with a common production method based on different biometric characteristics. For example, a set of images, all printed on the same kind of paper but corresponding to different faces, constitutes one single PAI species.

PAI species present a source of systematic variation in a test and can have significantly different error rates, making it difficult or even impossible to have a complete model of all possible presentation attack instruments. Furthermore, for the same PAI species, there will be a random variation between the results obtained by each specific instrument. Within each PAI species, the uncertainty associated with estimating the PAD error rate will depend on the number of artefacts tested and the number of biometric characteristics (e.g. individuals when testing an FRS-PAD). Due to the above, it cannot
be assumed that the error rates measured from one set of PAI are applicable to a completely different set.

The performance of each PAD mechanism shall be measured according to ISO/IEC 30107-3 \footnote{\url{https://www.iso.org/standard/79520.html}} using the Bona fide Presentation Classification Error Rate (BPCER), and Attack Presentation Classification Error Rate (APCER) metric defined as (\ref{eq:bpcer}) and (\ref{eq:apcer}).

\begin{equation}\label{eq:bpcer}
    BPCER=\frac{\sum_{i=1}^{N_{BF}}RES_{i}}{N_{BF}}
\end{equation}

\begin{equation}\label{eq:apcer}
    APCER=\frac{1}{N_{PAIS}}\sum_{i=1}^{N_{PAIS}}(1-RES_{i})
\end{equation}

where $N_{BF}$ is the number of bona fide presentations, $N_{PAIS}$ is the number of presentation attacks for a given presentation species and $RES_{i}$ is $1$ if the system's response to the $i-th$ attack is classified as an attack and $0$ if classified as bona fide. 
A PAD performance is reported using the Equal Error Rate (EER), which is the point where the APCER is equal to BPCER. Also, two operational points are reported BPCER10 and BPCER20. The BPCER20 is the BPCER value obtained when the APCER is fixed at 5\%, and BPCER10 (APCER at 10\%).
\vspace{-0.3cm}

\section{Test methodology}
\label{sec:method}
\vspace{-0.1cm}

\subsection{Tests and protocols}

One of the main challenges of PAD algorithms is their generalisation capabilities under intra-dataset, cross-dataset and leave-one-out scenarios. In order to know these capabilities and the influences of our new dataset Flirck-PAD, we included it as a train and test set separately. The following test protocols are defined.

\subsubsection*{Intra-dataset testing}

This protocol describes one of the most widely used datasets to assess the discrimination ability of the PAD models. The training and test data come from the same datasets and share a similar domain distribution in terms of setting, lighting conditions, attack types, dynamic behaviour of the test subject, etc. It generally consists of training the algorithms with images/videos that are subject-disjoint with the test set but captured under very similar conditions.

\subsubsection*{Cross-dataset testing}

This is one of the most challenging protocols, given that a PAD is tested on an unknown attack presentation in unseen conditions, such as differences in the wild conditions: lighting, capture devices, etc. Our new dataset (F) is used only as a test set in this protocol.

\subsubsection*{Leave-One-Out testing}

LOO protocol, as we refer to in this paper, is used to evaluate our proposed method in a challenging environment to represent the real deployment conditions. The protocol was applied considering four datasets and also our new database Flickr-PAD. Then in total, we evaluated five datasets. We used eight LOO protocols in this work, combining the datasets C, M, O, R, and F described in Section \ref{sec:database}.

First, four databases for training and validation were selected, and one remaining dataset (left out) to test the model as an unknown scenario, e.g. train with Casia(C), plus MSU(M) plus Oulu-NPU(O) plus Flickr-PAD (F) and evaluate with Replay-attack (R). Iteratively, to cover all datasets, we interchange the datasets to include the left-out dataset in the train/validation intra-dataset test steps and leave out one. Our new dataset (F) is used only as a training set in this protocol.

The distribution of each dataset for training, validation, intra-dataset and cross-dataset test is shown in Table \ref{distribution_dataset}.

\begin{table}[H]
\scriptsize
\caption{Dataset distribution for experiments.}
\label{distribution_dataset}

\centering{}%
\begin{tabular}{|c|c|c|c|c|c|}
\hline
Dataset & Train & Val & Intra-dataset & Cross-dataset  & Total\tabularnewline
\hline
\hline
CASIA-MFSD\cite{CASIA-MFSD} & 787 & 245 & 144 & 1,176 & 1,176 \tabularnewline
\hline
MSU-MFSD\cite{MSU-MFSD} & 293 & 84 & 43 & 420 & 420 \tabularnewline
\hline
OULU-NPU\cite{OULU-NPU} & 6,590 & 1,843 & 978 & 9,417 & 9,417 \tabularnewline
\hline
Replay-Mobile\cite{replay-mobile} & 4,987 & 1,442 & 711 & 7,140 & 7,140 \tabularnewline
\hline
Flickr-PAD (\textbf{ours}) & 9,800 & 2,800 & 1,400 & 14,000 & 14,000 \tabularnewline
\hline
\end{tabular}
\end{table}

\subsection{Deep Learning Models}

Three PAD algorithms were implemented on PyTorch using the Kedro framework \cite{Alam_Kedro_2021}, MobileNet-V3-small, MobileNet-V3-large \cite{mobilenetv3}, and EfficientNet-B0 \cite{efficientnet}. The input image size for all algorithms was $224\times224\times3$ pixels.
The same Data Augmentation (DA) scheme based on the flip, coarse ($p=0.5$), rotation (10\textdegree), illumination changes and other operations were applied on all three algorithms based on the default parameters of albumentation library \cite{info11020125}. 


The criteria for selecting these networks was to get lightweight models for a real operation.  Both architectures used the pre-trained weights from Imagenet. 
We modified the last layer of each net to be a two-class output instead of the original $1,000$ classes. The models were trained with an SGD optimiser with a momentum of $0.9$, a learning rate of $5e^{-4}$, with Cross-Entropy-Loss. The number of workers and batch size were both set to $32$. The training was done in $100$ epochs. All the hyperparameters were selected using a grid search.

\begin{table*}
\scriptsize
\caption{Summary results of tree protocols. The best results for each protocol and model are highlighted in colours.}
\label{results}
\scriptsize
\centering{}%
\begin{tabular}{|c|c|c|ccc|ccc|}
\hline
& & & \multicolumn{3}{c|}{Intra-dataset test} & \multicolumn{3}{c|}{Cross-dataset test}\tabularnewline
\hline
Model & Experiment & Protocol & \multicolumn{1}{c|}{$EER${[}\%{]}} & \multicolumn{1}{c|}{$BPCER_{10}${[}\%{]}} & $BPCER_{20}${[}\%{]} & \multicolumn{1}{c|}{$EER${[}\%{]}} & \multicolumn{1}{c|}{$BPCER_{10}${[}\%{]}} & $BPCER_{20}${[}\%{]}\tabularnewline
\hline
\hline
\multirow{12}{*}{MobileNet-V3 small} & \multirow{4}{*}{1} & CMO-R  & \multicolumn{1}{c|}{0.00} & \multicolumn{1}{c|}{0.00} & 0.00 & \multicolumn{1}{c|}{25.72} & \multicolumn{1}{c|}{41.00} & 53.01\tabularnewline
\cline{3-9} 
& & RMO-C & \multicolumn{1}{c|}{0.00} & \multicolumn{1}{c|}{0.00} & 0.00 & \multicolumn{1}{c|}{43.37} & \multicolumn{1}{c|}{84.59} & 89.61\tabularnewline
\cline{3-9} 
& & CRO-M & \multicolumn{1}{c|}{0.00} & \multicolumn{1}{c|}{0.00} & 0.00 & \multicolumn{1}{c|}{43.81} & \multicolumn{1}{c|}{75.24} & 87.62\tabularnewline
\cline{3-9} 
& & CMR-O & \multicolumn{1}{c|}{0.00} & \multicolumn{1}{c|}{0.00} & 0.00 & \multicolumn{1}{c|}{24.15} & \multicolumn{1}{c|}{48.35} & 63.05\tabularnewline
\cline{2-9} 
& \multirow{4}{*}{2} & CMO-F & \multicolumn{1}{c|}{-} & \multicolumn{1}{c|}{-} & - & \multicolumn{1}{c|}{26.73} & \multicolumn{1}{c|}{52.52} & 65.46\tabularnewline
\cline{3-9} 
& & RMO-F & \multicolumn{1}{c|}{-} & \multicolumn{1}{c|}{-} & - & \multicolumn{1}{c|}{31.53} & \multicolumn{1}{c|}{62.34} & 76.40\tabularnewline
\cline{3-9} 
& & CRO-F & \multicolumn{1}{c|}{-} & \multicolumn{1}{c|}{-} & - & \multicolumn{1}{c|}{27.85} & \multicolumn{1}{c|}{55.36} & 71.09\tabularnewline
\cline{3-9} 
& & CMR-F & \multicolumn{1}{c|}{-} & \multicolumn{1}{c|}{-} & - & \multicolumn{1}{c|}{30.80} & \multicolumn{1}{c|}{63.13} & 76.67\tabularnewline
\cline{2-9} 
& \multirow{4}{*}{3} & CMOF-R & \multicolumn{1}{c|}{0.52} & \multicolumn{1}{c|}{0.00} & 0.00 & \multicolumn{1}{c|}{\textbf{9.47}} & \multicolumn{1}{c|}{\textbf{8.91}} & \textbf{13.87}\tabularnewline
\cline{3-9} 
& & RMOF-C & \multicolumn{1}{c|}{0.13} & \multicolumn{1}{c|}{0.00} & 0.00 & \multicolumn{1}{c|}{\textbf{27.96}} & \multicolumn{1}{c|}{\textbf{57.71}} & \textbf{68.10}\tabularnewline
\cline{3-9} 
& & CROF-M & \multicolumn{1}{c|}{0.45} & \multicolumn{1}{c|}{0.00} & 0.00 & \multicolumn{1}{c|}{\textbf{19.05}} & \multicolumn{1}{c|}{\textbf{24.76}} & \textbf{39.05}\tabularnewline
\cline{3-9} 
& & CMRF-O & \multicolumn{1}{c|}{0.32} & \multicolumn{1}{c|}{0.00} & 0.00 & \multicolumn{1}{c|}{\textbf{17.19}} & \multicolumn{1}{c|}{\textbf{42.03}} & \textbf{61.96}\tabularnewline
\hline
\hline
\multirow{12}{*}{MobileNet-V3 large} & \multirow{4}{*}{1} & CMO-R & \multicolumn{1}{c|}{0.00} & \multicolumn{1}{c|}{0.00} & 0.00 & \multicolumn{1}{c|}{17.09} & \multicolumn{1}{c|}{26.81} & 34.59\tabularnewline
\cline{3-9} 
& & RMO-C & \multicolumn{1}{c|}{0.00} & \multicolumn{1}{c|}{0.00} & 0.00 & \multicolumn{1}{c|}{34.72} & \multicolumn{1}{c|}{67.38} & 77.79\tabularnewline
\cline{3-9} 
& & CRO-M & \multicolumn{1}{c|}{0.00} & \multicolumn{1}{c|}{0.00} & 0.00 & \multicolumn{1}{c|}{46.67} & \multicolumn{1}{c|}{72.38} & 81.90\tabularnewline
\cline{3-9} 
& & CMR-O & \multicolumn{1}{c|}{0.00} & \multicolumn{1}{c|}{0.00} & 0.00 & \multicolumn{1}{c|}{23.24} & \multicolumn{1}{c|}{41.88} & 74.50\tabularnewline
\cline{2-9} 
& \multirow{4}{*}{2} & CMO-F & \multicolumn{1}{c|}{-} & \multicolumn{1}{c|}{-} & - & \multicolumn{1}{c|}{22.74} & \multicolumn{1}{c|}{40.71} & 54.83\tabularnewline
\cline{3-9} 
& & RMO-F & \multicolumn{1}{c|}{-} & \multicolumn{1}{c|}{-} & - & \multicolumn{1}{c|}{25.61} & \multicolumn{1}{c|}{48.63} & 62.70\tabularnewline
\cline{3-9} 
& & CRO-F & \multicolumn{1}{c|}{-} & \multicolumn{1}{c|}{-} & - & \multicolumn{1}{c|}{18.44} & \multicolumn{1}{c|}{29.60} & 41.98\tabularnewline
\cline{3-9} 
& & CMR-F & \multicolumn{1}{c|}{-} & \multicolumn{1}{c|}{-} & - & \multicolumn{1}{c|}{23.69} & \multicolumn{1}{c|}{41.93} & 59.53\tabularnewline
\cline{2-9} 
& \multirow{4}{*}{3} & CMOF-R  & \multicolumn{1}{c|}{0.10} & \multicolumn{1}{c|}{0.00} & 0.00 & \multicolumn{1}{c|}{\textcolor{blue}{7.85}} & \multicolumn{1}{c|}{\textcolor{blue}{7.08}} & \textcolor{blue}{11.15}\tabularnewline
\cline{3-9} 
& & RMOF-C & \multicolumn{1}{c|}{0.00} & \multicolumn{1}{c|}{0.00} & 0.00 & \multicolumn{1}{c|}{\textcolor{blue}{23.61}} & \multicolumn{1}{c|}{\textcolor{blue}{41.58}} & \textcolor{blue}{50.90}\tabularnewline
\cline{3-9} 
& & CROF-M & \multicolumn{1}{c|}{0.10} & \multicolumn{1}{c|}{0.00} & 0.00 & \multicolumn{1}{c|}{\textcolor{blue}{19.05}} & \multicolumn{1}{c|}{\textcolor{blue}{23.81}} & \textcolor{blue}{28.57}\tabularnewline
\cline{3-9} 
& & CMRF-O & \multicolumn{1}{c|}{0.00} & \multicolumn{1}{c|}{0.00} & 0.00 & \multicolumn{1}{c|}{\textcolor{blue}{16.67}} & \multicolumn{1}{c|}{\textcolor{blue}{37.06}} & \textcolor{blue}{73.28}\tabularnewline
\hline
\hline
\multirow{12}{*}{EfficientNet B0} & \multirow{4}{*}{1} & CMO-R  & \multicolumn{1}{c|}{0.00} & \multicolumn{1}{c|}{0.00} & 0.00 & \multicolumn{1}{c|}{15.90} & \multicolumn{1}{c|}{19.15} & 27.88\tabularnewline
\cline{3-9} 
& & RMO-C & \multicolumn{1}{c|}{0.00} & \multicolumn{1}{c|}{0.00} & 0.00 & \multicolumn{1}{c|}{40.28} & \multicolumn{1}{c|}{76.34} & 86.38\tabularnewline
\cline{3-9} 
& & CRO-M & \multicolumn{1}{c|}{0.00} & \multicolumn{1}{c|}{0.00} & 0.00 & \multicolumn{1}{c|}{43.81} & \multicolumn{1}{c|}{69.52} & 71.43\tabularnewline
\cline{3-9} 
& & CMR-O & \multicolumn{1}{c|}{0.00} & \multicolumn{1}{c|}{0.00} & 0.00 & \multicolumn{1}{c|}{19.26} & \multicolumn{1}{c|}{37.33} & 51.73\tabularnewline
\cline{2-9} 
& \multirow{4}{*}{2} & CMO-F  & \multicolumn{1}{c|}{-} & \multicolumn{1}{c|}{-} & - & \multicolumn{1}{c|}{13.74} & \multicolumn{1}{c|}{19.13} & 29.14\tabularnewline
\cline{3-9} 
& & RMO-F & \multicolumn{1}{c|}{-} & \multicolumn{1}{c|}{-} & - & \multicolumn{1}{c|}{19.94} & \multicolumn{1}{c|}{36.18} & 52.98\tabularnewline
\cline{3-9} 
& & CRO-F & \multicolumn{1}{c|}{-} & \multicolumn{1}{c|}{-} & - & \multicolumn{1}{c|}{22.70} & \multicolumn{1}{c|}{43.93} & 59.72\tabularnewline
\cline{3-9} 
& & CMR-F & \multicolumn{1}{c|}{-} & \multicolumn{1}{c|}{-} & - & \multicolumn{1}{c|}{25.16} & \multicolumn{1}{c|}{50.20} & 66.08\tabularnewline
\cline{2-9} 
& \multirow{4}{*}{3} & CMOF-R  & \multicolumn{1}{c|}{0.19} & \multicolumn{1}{c|}{0.00} & 0.00 & \multicolumn{1}{c|}{\textcolor{orange}{9.99}} & \multicolumn{1}{c|}{\textcolor{orange}{9.94}} & \textcolor{orange}{13.28}\tabularnewline
\cline{3-9} 
& & RMOF-C & \multicolumn{1}{c|}{0.13} & \multicolumn{1}{c|}{0.00} & 0.00 & \multicolumn{1}{c|}{\textcolor{orange}{25.52}} & \multicolumn{1}{c|}{\textcolor{orange}{53.76}} & \textcolor{orange}{62.72}\tabularnewline
\cline{3-9} 
& & CROF-M & \multicolumn{1}{c|}{0.13} & \multicolumn{1}{c|}{0.00} & 0.00 & \multicolumn{1}{c|}{\textcolor{orange}{18.10}} & \multicolumn{1}{c|}{\textcolor{orange}{29.52}} & \textcolor{orange}{42.86}\tabularnewline
\cline{3-9} 
& & CMRF-O & \multicolumn{1}{c|}{0.16} & \multicolumn{1}{c|}{0.00} & 0.00 & \multicolumn{1}{c|}{\textcolor{orange}{13.11}} & \multicolumn{1}{c|}{\textcolor{orange}{18.57}} & \textcolor{orange}{35.82}\tabularnewline
\hline
\end{tabular}
\end{table*}

\section{EXPERIMENTS AND RESULTS}
\label{sec:exp_results}

In this paper, three different protocols were defined in order to evaluate the influence of our new proposed Flickr-PAD database. According to each experiment, we explored five cross-dataset tests.  In the end, we performed 12 evaluations for each protocol.
Results of all experiments are shown in Table \ref{results}. All the datasets were divided into sub-sets 70,0\% train, 20,0\% validation and 10,0\% for the test.

\subsection{Experiment 1: Baseline}
First, we trained our models using four state-of-the-art datasets as a baseline. Following our LOO protocol, we trained and validated the model with the fusion of three databases. We then  evaluated in an intra-dataset test (also from these three databases). The fourth remaining dataset is used as a test set in LOO. This means training with C, plus M, plus O and testing in R. Afterward, iteratively changed all the datasets as a one-out test set. The best performance obtained in this experiment was the model based on EfficientNet-B0 using the CMO-R protocol with EER of 15.90\%, BPCER10 of 19.15\%, and BPCER20 of 27.88\%. All models obtained with the CMO-R protocol returned the best BPCER10 and BPCER20 results, as shown in Table \ref{results}.

\subsection{Experiment 2: Baseline plus Flickr-PAD dataset on test}
The second experiment consisted of cross-dataset testing of the models obtained in Experiment 1 using the Flickr-PAD dataset as the test set. Again, EfficientNet-B0 achieved the best EER of 13.74\%, BPCER10 of 19.13\%, and BPCER20 of 29.14\%, with the CMO-F protocol. In general, the F test returned less error in comparison to C and M, but a similar performance was reached when compared with O and R.

\subsection{Experiment 3: Baseline augmented with Flickr-PAD database}

The third experiment consisted of training, validating, and intra-dataset testing with three state-of-the-art datasets plus the Flickr-PAD dataset and cross-dataset testing with the remaining state-of-the-art dataset (skipping F) following the LOO protocol. In this experiment, our Flirck-PAD was used only to augment the training set. 
MobileNet-V3-small showed the best improvement when comparing Experiment 1 to Experiment 3. In protocol CMOF-R, cross-dataset EER dropped from 25.72\% to 9.47\%, BPCER10 from 41.00\% to 8.91\%, and BPCER20 from 53.01\% to 13.87\%, as shown in Table \ref{results}.

In protocols RMO-C and CRO-M, EfficientNet-B0 presented the best improvement after adding F to the train set. In Table \ref{results}, EER dropped from 43.81\% to 18.10\%, BPCER10 from 69.52\% to 29.52\%, and BPCER20 from 71.43\% to 42.86\%, in CRO-M protocol.

The models based on MobileNet-V3-large net and trained with the CMOF-R protocol showed the best result, with EER of 7.85\%, BPCER10 of 7.08\%, and BPCER20 of 11.15\%. Every model improved its performance when adding F to the train and validation.

\begin{figure*}
\centering
\includegraphics[scale=0.17]{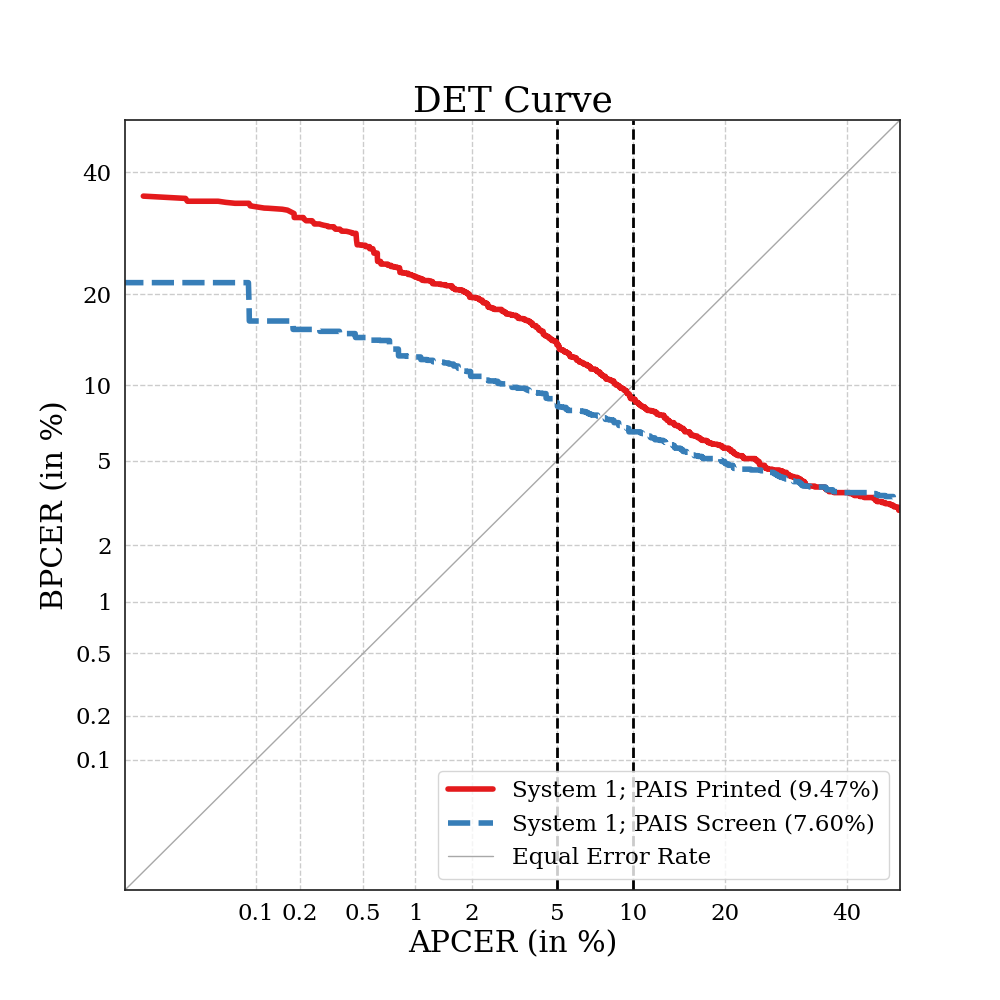}
\includegraphics[scale=0.17]{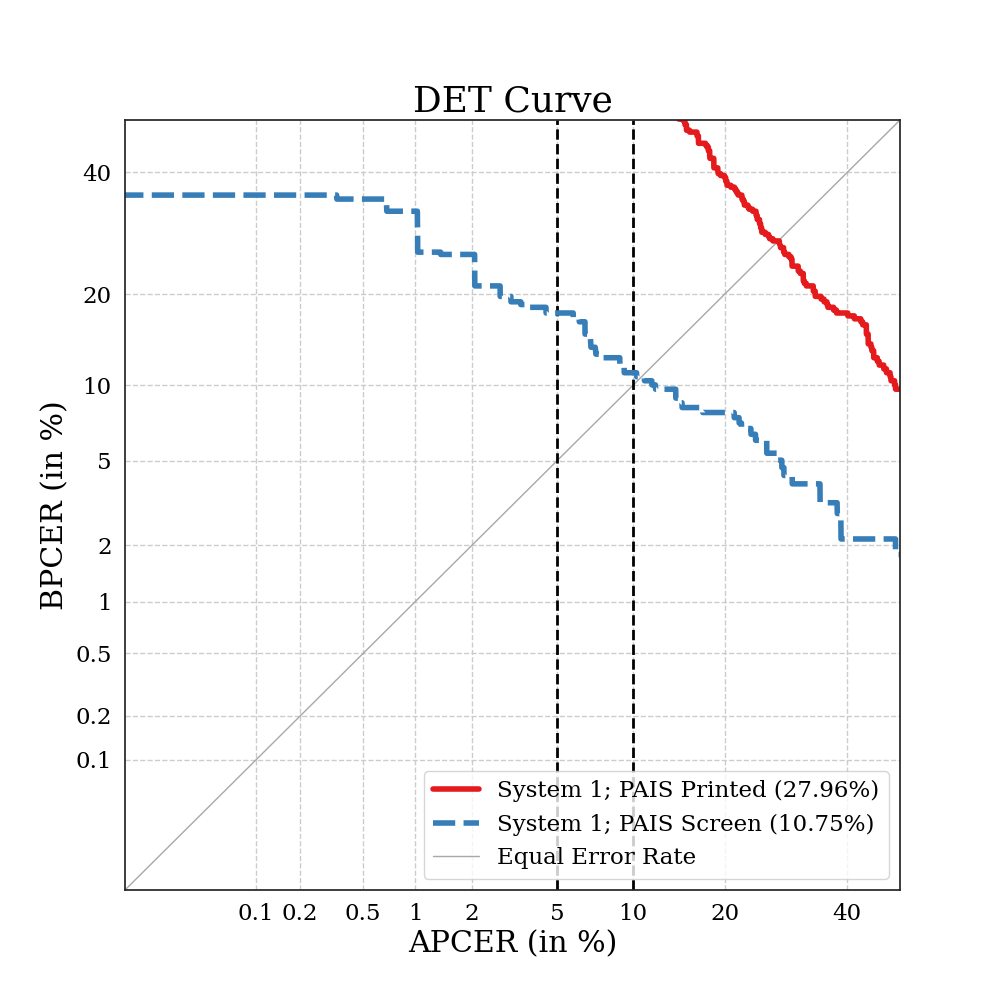}
\includegraphics[scale=0.17]{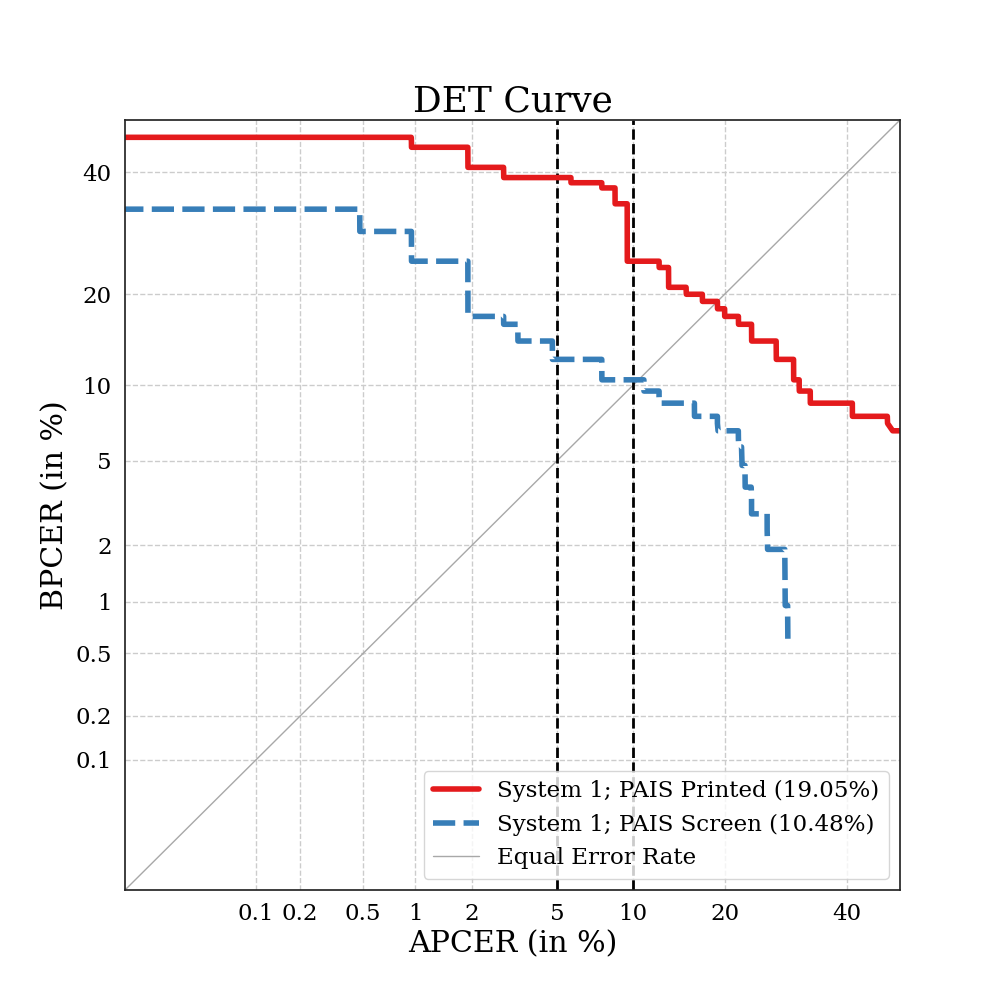}
\includegraphics[scale=0.17]{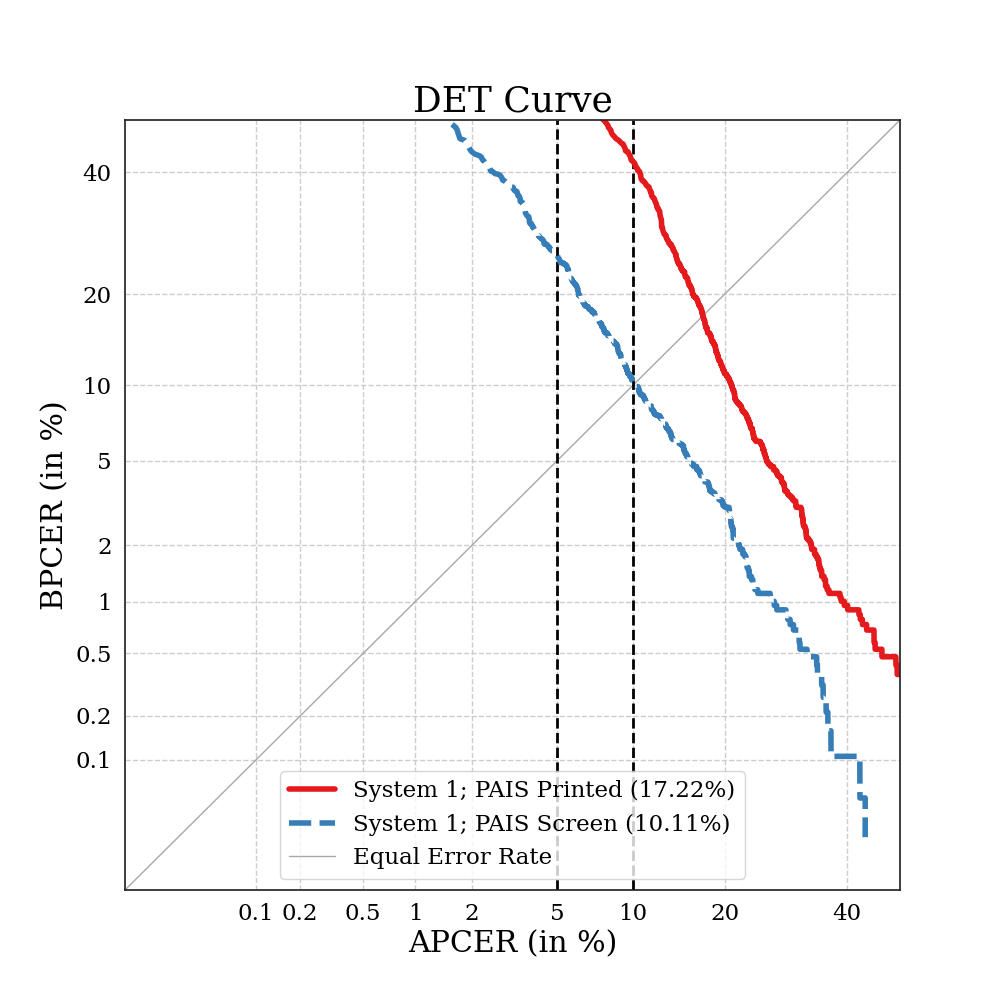}
\caption{\label{DETcurves_mobilenetv3small_w:wout_flickr} Cross-dataset test DET Curves obtained with MobileNet-V3 small. Experiment results (training with LOO iteratively by CMR and testing with F dataset). Left to right: Train with CMO, RMO, CRO and CMR protocols. Dot-line indicates BPCER10 and BPCER20, respectively. The EER is reported in parentheses.}
\end{figure*}

Figure \ref{DETcurves_mobilenetv3small_w:wout_flickr} shows all the DET curves from models based on MobileNet-V3-small result when Flirck-PAD is used in the train set.

\vspace{-0.3cm}

\section{Conclusions}
\label{sec:conclusions}

This paper evaluated the performance of three CNN models (MobileNetV3 small and large and EfficientNet-B0) at PAD using four SOTA datasets and our new PAD database: Flickr-PAD. Then, more realistic operational results for remote verification can be obtained based on images captured in high resolution as usual nowadays.
Overall, our new dataset was shown to be challenging when used as a test set (Experiment 2). When used in training, adding our new dataset helps to improve the results, as is shown in Experiment 3. The lower EER reached was 9.47\% using MobileNet-V3-small and 7.85 \%  when using V3-large. 





\bibliographystyle{myieee}
\bibliography{biblio}

\begin{thebibliography}{10}\itemsep=-1pt

\bibitem{yu2022deep}
Z.~Yu, Y.~Qin, X.~Li, C.~Zhao, Z.~Lei, and G.~Zhao.
\newblock Deep learning for face anti-spoofing: A survey.
\newblock {\em IEEE Trans. on Pattern Analysis and Machine Intelligence
  (TPAMI)}, 2022.

\bibitem{8987257}
S.~Mehta, A.~Uberoi, A.~Agarwal, M.~Vatsa, and R.~Singh.
\newblock Crafting a panoptic face presentation attack detector.
\newblock In {\em 2019 Intl. Conf. on Biometrics (ICB)}, pages 1--6, 2019.

\bibitem{8955089}
P.~K. Das, B.~Hu, C.~Liu, K.~Cui, P.~Ranjan, and G.~Xiong.
\newblock A new approach for face anti-spoofing using handcrafted and deep
  network features.
\newblock In {\em 2019 IEEE Intl. Conf. on Service Operations and Logistics,
  and Informatics (SOLI)}, pages 33--38, 2019.

\bibitem{8737949}
H.~Chen, G.~Hu, Z.~Lei, Y.~Chen, N.~M. Robertson, and S.~Z. Li.
\newblock Attention-based two-stream convolutional networks for face spoofing
  detection.
\newblock {\em IEEE Trans. on Information Forensics and Security}, 15:578--593,
  2020.

\bibitem{CASIA-MFSD}
Z.~Zhang, J.~Yan, S.~Liu, Z.~Lei, D.~Yi, and S.~Z. Li.
\newblock A face antispoofing database with diverse attacks.
\newblock In {\em 2012 5th IAPR Intl. Conf. on Biometrics (ICB)}, pages 26--31,
  2012.

\bibitem{OULU-NPU}
Z.~Boulkenafet, J.~Komulainen, L.~Li, X.~Feng, and A.~Hadid.
\newblock Oulu-npu: A mobile face presentation attack database with real-world
  variations.
\newblock In {\em 2017 12th IEEE Intl. Conf. on Automatic Face Gesture
  Recognition (FG 2017)}, pages 612--618, 2017.

\bibitem{replay-attack}
I.~Chingovska, A.~Anjos, and S.~Marcel.
\newblock On the effectiveness of local binary patterns in face anti-spoofing.
\newblock In {\em BIOSIG - Proc. of the Intl. Conf. of Biometrics Special
  Interest Group (BIOSIG)}, pages 1--7, 2012.

\bibitem{Jia_2020_CVPR_SSDG}
Y.~Jia, J.~Zhang, S.~Shan, and X.~Chen.
\newblock Single-side domain generalization for face anti-spoofing.
\newblock In {\em Proc. IEEE Conf. on Computer Vision and Pattern Recognition
  (CVPR)}, 2020.

\bibitem{fang2021learnable}
M.~Fang, N.~Damer, F.~Kirchbuchner, and A.~Kuijper.
\newblock Learnable multi-level frequency decomposition and hierarchical
  attention mechanism for generalized face presentation attack detection.
\newblock In {\em IEEE/CVF Winter Conf. on Appl. of Computer Vision (WACV)},
  pages 1131--1140, 2022.

\bibitem{Ebihara}
A.~F. Ebihara, K.~Sakurai, and H.~Imaoka.
\newblock Efficient face spoofing detection with flash.
\newblock {\em IEEE Trans. on Biometrics, Behavior, and Identity Science},
  3(4):535--549, 2021.

\bibitem{MSU-MFSD}
D.~Wen, H.~Han, and A.~K. Jain.
\newblock Face spoof detection with image distortion analysis.
\newblock {\em IEEE Trans. on Information Forensics and Security},
  10(4):746--761, 2015.

\bibitem{replay-mobile}
A.~Costa-Pazo, S.~Bhattacharjee, E.~Vazquez-Fernandez, and S.~Marcel.
\newblock The replay-mobile face presentation-attack database.
\newblock In {\em Intl. Conf. of the Biometrics Special Interest Group
  (BIOSIG)}, pages 1--7, 2016.

\bibitem{FlickrPaper}
T.~Karras, S.~Laine, and T.~Aila.
\newblock A style-based generator architecture for generative adversarial
  networks.
\newblock In {\em IEEE/CVF Conf. on Computer Vision and Pattern Recognition
  (CVPR)}, pages 4396--4405, 2019.

\bibitem{Alam_Kedro_2021}
S.~Alam, L.~Balan, G.~Comym, Y.~Dada, I.~Danov, L.~Hoang, R.~Kanchwala,
  J.~Klein, A.~Milne, J.~Schwarzmann, M.~Theisen, and S.~Wong.
\newblock {Kedro}, 12 2021.

\bibitem{mobilenetv3}
A.~Howard, M.~Sandler, B.~Chen, W.~Wang, L.-C. Chen, M.~Tan, G.~Chu,
  V.~Vasudevan, Y.~Zhu, R.~Pang, H.~Adam, and Q.~Le.
\newblock Searching for mobilenetv3.
\newblock In {\em 2019 IEEE/CVF Intl. Conf. on Computer Vision (ICCV)}, pages
  1314--1324, 2019.

\bibitem{efficientnet}
M.~Tan and Q.~Le.
\newblock {E}fficient{N}et: Rethinking model scaling for convolutional neural
  networks.
\newblock In K.~Chaudhuri and R.~Salakhutdinov, editors, {\em Proc. of the 36th
  Intl. Conf. on Mach. Lear.}, volume~97, pages 6105--6114, 2019.

\bibitem{info11020125}
A.~Buslaev, V.~I. Iglovikov, E.~Khvedchenya, A.~Parinov, M.~Druzhinin, and
  A.~A. Kalinin.
\newblock Albumentations: Fast and flexible image augmentations.
\newblock {\em Information}, 11(2), 2020.

\end{thebibliography}

\end{document}